\documentclass[runningheads]{llncs}
\usepackage{makeidx}  
\usepackage{mathtools}
\usepackage{enumitem}
\usepackage{hyperref}
\usepackage[capitalise]{cleveref}
\usepackage{amsfonts}
\usepackage{algorithm}
\usepackage{algpseudocode}
\usepackage{epsfig}
\usepackage{svg}
\usepackage{amsmath}
\usepackage{hyperref}
\usepackage{tabularx}
\usepackage{verbatim}
\usepackage{booktabs}
\usepackage[scaled]{helvet}

\begin{document}
%
%
\title{Action-Evolution Petri Nets: a Framework for Modeling and Solving Dynamic Task Assignment Problems}
\titlerunning{Action-Evolution Petri Nets}  
%
\author{Riccardo Lo Bianco\inst{1} \and
Remco Dijkman\inst{1} \and
Wim Nuijten\inst{1, 2} \and
Willem van Jaarsveld\inst{1}\
}

\authorrunning{Riccardo Lo Bianco et al.}   
%
\tocauthor{Riccardo Lo Bianco}
\institute{$^1$ Eindhoven University of Technology, Netherlands\newline
\email{\{r.lo.bianco|r.m.dijkman|w.p.m.nuijten|w.l.v.jaarsveld\}@tue.nl}\newline
$^2$ Eindhoven Artificial Intelligence Systems Institute, Netherlands\newline
}

\maketitle
%
\begin{abstract}
Dynamic task assignment involves assigning arriving tasks to a limited number of resources in order to minimize the overall cost of the assignments. To achieve optimal task assignment, it is necessary to model the assignment problem first. While there exist separate formalisms, specifically Markov Decision Processes and (Colored) Petri Nets, to model, execute, and solve different aspects of the problem, there is no integrated modeling technique. To address this gap, this paper proposes Action-Evolution Petri Nets (A-E PN) as a framework for modeling and solving dynamic task assignment problems. A-E PN provides a unified modeling technique that can represent all elements of dynamic task assignment problems. Moreover, A-E PN models are executable, which means they can be used to learn close-to-optimal assignment policies through Reinforcement Learning (RL) without additional modeling effort. To evaluate the framework, we define a taxonomy of archetypical assignment problems. We show for three cases that A-E PN can be used to learn close-to-optimal assignment policies. Our results suggest that A-E PN can be used to model and solve a broad range of dynamic task assignment problems.
\end{abstract}

\section{Introduction}

During the execution of a business process, tasks become executable and resources become available to execute these tasks. As resources are assigned to tasks, they become unavailable to execute other tasks. Consequently, continuously assigning the right task to the right resource is essential to run a process efficiently. This problem is known as dynamic task assignment. The dynamic task assignment problem can be seen as a particular case of the \textit{dynamic assignment problem}, which, according to~\cite{kuhn_hungarian_1955}, is the problem of assigning a fixed number of individuals to a sequence of tasks, such as to minimize the total cost of the allocations, which may include setup costs, travel costs, or other time-varying costs. This problem has been extensively studied in business process optimization~\cite{gulpinar_heuristics_2018} as well as related areas, such as manufacturing~\cite{hu_petri-net-based_2020}. For the sake of brevity, we will employ the term ``assignment problem'' to indicate the general dynamic (task) assignment problem.

To solve an assignment problem, it must first be modeled mathematically. Markov Decision Processes (MDPs) are a common technique for modeling assignment problems~~\cite{spivey_dynamic_2004}, and they are the standard interface for Reinforcement Learning (RL) algorithms~\cite{sutton_reinforcement_nodate}. The basic definition of MDP involves a single agent interacting with an environment to maximize a cumulative reward, which is a global signal of the goodness of the actions chosen by the agent during a (possibly infinite) sequence of system states. In the context of business process optimization, the environment is the business process that must be executed, and the agent decides which task to assign to which resource. The reward is calculated based on what we want to optimize in the process, such as the total time resources spend working, the total cost of employing the resources, or the time customers spend waiting. While MDPs provide a good formalism for modeling the agent's behavior, they consider the environment, in our case the business process, as a black box that provides rewards for the decisions taken by the agent without exposing its internal behavior. Moreover, they do not have an agreed-upon syntax and lack any type of graphical representation. On the other hand, (Colored) Petri Nets~\cite{goos_brief_1997} are a well-known formalism for modeling a business process but have no inherent mechanisms for modeling and calculating the best decision in a given situation. Also, frameworks exist for many mathematical optimization techniques, such as linear programming and constraint programming, where problems can be modeled and solved without additional effort. However, no such framework exists for dynamic task assignment problems.

To fill this gap, this paper presents a unified and executable framework for modeling assignment problems. We use the term ``unified'' to refer to the capability of expressing both the agent and the environment of the assignment problem in a single standardized notation, thus simplifying the modeling of new problems. We use the term ``executable'' to refer to the possibility of using the models to train and test decision-making algorithms (specifically RL algorithms) without additional effort. To this end, we propose a new artifact in the form of a modeling language with a solid mathematical foundation, namely A-E Petri Net (A-E PN), which draws from the well-known Petri Net (PN) formalism to model assignment problems in a readable and executable manner. This paper pays particular attention to embedding the A-E PN formalism in the RL cycle, such that RL algorithms can be trained and used to solve assignment problems without additional effort.

The proposed artifact is evaluated by modeling and solving a set of archetypical assignment problems. A taxonomy of assignment problem variants is proposed, and an example for each of the three main variants is modeled through A-E PN. An RL algorithm is trained on each instance, achieving close-to-optimal results. Apart from modeling each assignment problem as an A-E PN, no additional effort is required to achieve these results, empirically demonstrating that A-E PN constitutes a unified and executable framework for modeling and solving assignment problems.

Against this background, the remainder of this paper is structured as follows. Section \ref{related_work} is dedicated to a review of relevant literature. Section \ref{preliminaries} introduces Timed-Arc Colored Petri Nets (T-A CPN). Section \ref{ae_pn} is devoted to the formal definition of Action-Evolution Petri Net and the description of the integration of A-E PN in the classic RL loop. In section \ref{use_cases}, an essential taxonomy of assignment problem variants is presented. A problem instance for each variant is modeled through A-E PN, and a RL algorithm is trained on each instance, obtaining close-to-optimal results. Section \ref{conclusion} discusses the proposed method's benefits and limitations and delineates the next research steps.  

\section{Related work}\label{related_work}

To the best of our knowledge, this paper presents the first attempt at defining a unified and executable framework for assignment problems. In contrast, the relation between (generalized stochastic) Petri Nets and Markov Chains is well studied~\cite{bause_stochastic_1998}, but Markov Chains cannot be used to model and optimize (task assignment) decisions. Since Markov Decision Processes can be seen as an extension to Markov Chains, the idea of extending Petri Nets to model Markov Decision Processes follows naturally. Several attempts at this exist in the literature, but none focus on the assignment problem. An overview of existing frameworks for modeling and solving dynamic optimization problems is presented in \cref{tab:literature}, listing, for each framework, the Petri Net variant employed, the scope of applicability, and whether the framework is unified and executable. The current work is presented in the last line.

\setlength{\tabcolsep}{2pt}
\begin{table}[]
\caption{Comparison of existing frameworks for dynamic optimization.}
\begin{tabular}{lllll}
\toprule

                                                                            Reference     & PN & Scope                                                                      & Unified & Executable             \\ \midrule
\multicolumn{1}{l}{\cite{da_rocha_costa_markov_2010}}     & FPN        & Problems expressible as finite MDPs                                        & Yes     & Yes* \\
\multicolumn{1}{l}{\cite{hutchison_markov_2007}}      & DPN        & Problems expressible as finite MDPs                                        & No      & Yes                    \\
\multicolumn{1}{l}{\cite{qiu_dynamic_2000}}           & GSPN       & A single power management problem                                  & Yes     & No                     \\ 
\multicolumn{1}{l}{\cite{drakaki_manufacturing_2017}} & TCPN       & A single manufacturing scheduling problem                                  & Yes     & No                     \\ 
\multicolumn{1}{l}{\cite{riedmann_timed_2022}}        & TCPN       & Manufacturing scheduling problems  & Yes    & No                 \\ 
\multicolumn{1}{l}{This paper}                                                       & A-E PN     & Assignment problems                & Yes     & Yes                    \\ \bottomrule
\end{tabular}
\label{tab:literature}
\footnotesize{* No executable example is provided.}
\end{table}

In~\cite{da_rocha_costa_markov_2010}, the authors define a CPN variant: Factored Petri Net (FPN). In FPNs, the transition probabilities are defined explicitly, and a reward is attached to each network state. A limitation of~\cite{da_rocha_costa_markov_2010} is that actions must be input marks from a single source transition (a transition without input arcs), while our framework allows actions to be defined anywhere in the Petri net, thus allowing for more modeling flexiblity.

In~\cite{hutchison_markov_2007}, the authors propose the Decision Petri Net (DPN) formalism. In DPN, the network is partitioned into a probabilistic network, in which transition probabilities are determined on arcs, and a non-deterministic network, corresponding to the actions that can be taken at a given moment by the decision maker. In our framework, we remove the need for two separate subnets and model the agents as tokens in the network, obtaining a unified representation. Both~\cite{da_rocha_costa_markov_2010}, and~\cite{hutchison_markov_2007} require the number of states in the system to be finite, whereas our approach does not rely on states enumeration.

In~\cite{qiu_dynamic_2000}, the authors propose a model for a power-managed distributed computing system that is based on the Generalized Stochastic Petri Net (GSPN) formalism and provide a translation to the equivalent continuous-time MDP. The work demonstrates the expressive power of PN variants, but the resulting model is not executable. Also, the paper presents a single case study, while our approach is demonstrated to be generally applicable to modeling and solving problems with different characteristics.

In~\cite{drakaki_manufacturing_2017}, a manufacturing scheduling problem is modeled using Timed Colored Petri Nets (TCPN). The search for an optimal policy is implemented using Q-learning, where each action corresponds to a complete schedule, which is a path from the initial marking to a final marking of the TCPN representing the system, whereas in our case, an action corresponds to a single assignment, which allows for more flexible modeling of decisions. Moreover,~\cite{drakaki_manufacturing_2017} only covers a single case study, relying heavily on problem-specific heuristics. 

In~\cite{riedmann_timed_2022}, the authors provide an example usage of TCPN in the context of manufacturing systems, focusing on reinforcement learning as solving approach. While~\cite{riedmann_timed_2022} highlights the relationship between TCPN and RL, TCPNs are used only to describe the environment and not to train or test solving algorithms. In contrast, our work provides a unified and executable framework.

\section{Preliminaries}\label{preliminaries}
This section provides the formal definition of Colored Petri Net (CPN) and Timed-Arc Colored Petri Net (T-A CPN), which will be used to define the new formalism.

Colored Petri Net (CPN)~\cite{goos_brief_1997} is an extension of Petri Nets (PN) in which tokens have different characteristics called colors. In the remainder of this section, we rely on the CPN definition provided in~\cite{jensen_high-level_1991}.

\begin{definition}[Colored Petri Net]\label{def:cpn} A CPN is defined as a tuple \(CPN = ( \mathcal{E}, P, T, F, C, G, E, I)\), such that:

\begin{itemize}
\item \(\mathcal{E}\) is a finite set of types called \textit{color sets}. Each color set must be finite and non-empty.
\item \(P\) is a finite set of \textit{places}.
\item \(T\) is a finite set of \textit{transitions}, such that $P \cap T = \emptyset$
\item \(F \subseteq P \times T \cup T \times P\) is a finite set of \textit{arcs}.
\item \(C : P \to \mathcal{E}\) is a \textit{color function} that maps each place \(p\) into a set of possible token colors. Each token on \(p\) must have a color that belongs to the type $C(p)$, which is called the place's \textit{color set}.
\item \(G\) is a \textit{guard function}. It is defined from \(T\) into expressions such that for each \( t \in T\), $G(t)$ is a Boolean expression and \(Type(Var(G(t))) \subseteq \mathcal{E}\), where \(Type(x)\) denotes the type of $x$ and \(Var(f)\) denotes the set of free variables in the function $f$.
\item \(E\) is an arc expression function. It is defined from \(F\) into expressions such that for each \(f \in F\), \(Type(E(f)) = C(P(f))_{MS}\) and \(Type(Var(E(f))) \subseteq \mathcal{E}\) where \(P(f)\) is the place of \(f\). This means that each evaluation of the arc expression must yield a multi-set (indicated by the \(MS\) subscript) over the color set attached to the corresponding place.
\item \(I\) is an \textit{initialization function}. It is defined from \(P\) into expressions such that \(\forall p \in P: Type(I(p)) = C(p)_{MS}\). The initialization function determines the network's \textit{initial marking}.
\end{itemize}
\end{definition}

\begin{definition}[Marking] A marking of a CPN is a function $M$, such that for each place $p \in P$, it defines a multi-set of colors $C(p) \rightarrow \mathbb{N}$, which maps each possible color of the place to the number of times it occurs.
\end{definition}

For a place $p$ with colors $C(p) = \{c_1, c_2\}$, we also write $M(p) = c_1^n c_2^m$ to denote that $p$ has \(n)\) token with color $c_1$ and \(m\) tokens with color $c_2$. Since a marking is a multi-set, multi-set operations, such as $\geq$, $+$, and $-$, are available on markings.

\begin{sloppypar}
\begin{definition}[Binding] For a transition $t$, the variables $Var(t) = Var(G(t)) \cup \{Var(E(f)) | f \in F, T(f) = t\}$ represent the set of variables from the guard function and the expressions on its arcs, where \(T(f)\) is the transition of arc \(f\).

A binding of a transition $t \in T$ is a function $Y$ that maps each $v \in Var(t)$ to a color, such that $\forall v \in Var(v): Y(v) \in Type(v)$ and $G(t)\langle Y\rangle$ evaluates to true, where $f\langle Y\rangle$ denotes the evaluation of a function $f$ with its free variables bound as $Y$.
\end{definition}
\end{sloppypar}

For a transition $t$ with variables $Var(t) = \{v_1, v_2\}$, we also write $Y(t) = \langle v_1=c_1, v_2=c_2\rangle$ to denote that the binding $Y$ assigns color $c_1$ to variable $v_1$ and color $c_2$ to variable $v_2$.

We now define the behavior of a CPN through its firing rules.
\begin{definition}[CPN Firing Rules]\label{def:cpn_firing_rules}
\begin{enumerate}
\item A transition $t$ is enabled in marking $M$ for binding $Y$ if and only if $\forall (p, t) \in F: M(p) \geq E((p,t))\langle Y\rangle$.
\item An enabled transition can fire, changing the Marking $M$ into a marking $M'$, such that $\forall p \in P: M'(p) = M(p) - E((p,t))\langle Y\rangle + E((t, p))\langle Y\rangle$.
\end{enumerate}
\end{definition}

The standard CPN definition assumes that the effect of a firing is always instantaneous. To account for time, we will refer to a modified version of the Timed-Arc Petri Net (T-A PN) formulation~\cite{10.1007/978-3-642-18381-2_4}. Our version defines a global clock, updated according to a next-event time progression. This is also the time management paradigm implemented in CPN Tools~\cite{cpntools}, a widely adopted software for Petri Nets modeling.

\begin{definition}[Timed-Arc Colored Petri Net]\label{def:tacpn} A T-A CPN is defined by a tuple \(TACPN =( \mathcal{E}, P, T, F, C, G, E, I)\), where \(P, T, F, C, G, I\) are as in~\cref{def:cpn}, and \(\mathcal{E}\) and $E$ are adapted as follows:
\begin{itemize}
    \item \(\mathcal{E}\) is a finite set of timed types called \textit{timed color sets}. A color of a timed color set has both a value $v$ and a time $\tau$, we also denote this as $v@\tau$.
    \item \(E\) is an arc expression function. It is defined from \(F\) into tuples of two elements. For a given \(f\in F\), \(E(f)_0\) is defined the same as $E$ in ~\cref{def:cpn} and \(E(f)_1\) is a \textit{scalar increment}, thus \(\forall f \in F: Type(E(f)_1) = \mathbb{N}\), that indicates the generated tokens' time with reference to the global clock. The second tuple element is ignored for arcs outgoing from places and incoming to transitions since the scalar increment is only used when producing new tokens.
\end{itemize}
\end{definition}

Note that each color now has a time and consequently, each color in a marking and in a binding has time. For example, we can refer to the marking of a place $p$ with $M(p)=c_1@2^1c_1@3^5$ as the marking that has one token with color $c_1$ at time 2 and five tokens with color $c_1$ at time 3. With some abuse of notation, we will allow arc expression functions $E(f)_0$, to ignore the time element of colors and leave it unaffected, and we will denote with $c@e$ that an expression $e$ only changes the time element of a timed color.

We also extend the concept of marking to account for the presence of a global clock, which we need further on in the paper to define the transition rules for A-E PN.

\begin{definition}[Timed Marking]\label{def:tacpn_marking} A timed marking is defined as the tuple \(TM = (M, \tau)\), where \(M\) is a marking and \(\tau\) is the current value of the global clock.
\end{definition}

The T-A CPN firing rule can then be expressed as follows:

\begin{definition}[T-A CPN Firing Rules]\label{def:tacpn_firing_rules}
\begin{enumerate}
\item Let $t$ be a transition that is enabled in marking $M$ for binding $Y = \langle v1=c_1@\tau_1, v2=c_2@\tau_2, \ldots, v_n=c_n@\tau_n \rangle$ as in~\cref{def:cpn_firing_rules} (using only $E_0$ for $E$). The enabling time of the transition, denoted $\tau_E$, is $max(\tau_1, \tau_2, \ldots, \tau_n)$.
\item An enabled transition $t$ is time-enabled in timed marking $(M, \tau)$, if its enabling time $\tau_E$ is less than or equal to $\tau$, and there exists no transition $t'$ that is enabled in marking $M$ for some binding $Y'$ with enabling time $\tau'_E \leq \tau_E$.
\item A transition $t$ that is time-enabled in timed marking $(M, \tau)$ for binding $Y$ with enabling time $\tau_E$ can fire, changing the timed marking to $(M', \tau_E)$, where $M'$ is constructed, such that $\forall p \in P: M'(p) = M(p) - E((p,t))_0\langle Y\rangle + E((t, p))_0\langle Y\rangle@\tau_E+E((t, p))_1$.
\item When there exists no $t$ in timed marking $(M, \tau)$, for which there is a binding $Y$, such that $t$ is time-enabled, the global clock $\tau$ is increased until there is.
\end{enumerate}
\end{definition}
In practice, point 4 can be performed by evaluating bindings that are enabling but not time-enabling. The binding that leads to the lowest enabling time reveals the minimal increase of the global clock, making it possible to update the global clock using a next-event time progression.

\section{Action-Evolution Petri Nets}\label{ae_pn}

This section extends the definition of T-A CPN to provide a model that can automatically learn close-to-optimal task assignment policies. This extension is called Action-Evolution Petri Nets (A-E PN). The new elements are first described informally, then a formal definition is provided. Finally, the definition is incorporated into the RL cycle, allowing for automated learning of close-to-optimal task assignment policies.

\subsection{Tags and Rewards}\label{actions_evolutions}

The overall objective of A-E PN is to mimic the behavior of an agent that observes changes in the environment and acts upon those changes when possible. We will thus extend the CPN definition provided in the background section to distinguish two separate types of transitions:
\begin{itemize}
    \item \textbf{Actions}: transitions that represent actions taken by the agent. In the context of assignment problems, the firing of an action transition represents a single assignment.
    \item \textbf{Evolutions}: transitions that represent events happening in the system independently of the actions taken by the agent. The firing of an evolution transition represents a single event in the environment, for example, the arrival of a new order.
\end{itemize}

This distinction is expressed by associating every transition with a \textit{transition tag}, that can be either \(A\) (action) or \(E\) (evolution), through a \textit{transition tag function} \(L\). We also extend the concept of marking to embed a \textit{network tag} \(l\), which can assume a single value in \(\{A, E\}\): only transitions associated with a tag of the same type as the one in the network tag are allowed to fire.
The network tag \(l\) must be updated every time no transitions with the same tag are available for firing. The \textit{tag update function} \(S\) performs the update by changing the network's tag from $A$ to $E$ or vice versa: $S(l)=A \text{, if } l = E; S(l)=E \text{, if } l = A$. We use the term \textit{tag time frame} to refer to the period between changes in the network tag.

The objective of the RL cycle is the maximization of a cumulative reward over a (possibly infinite) horizon. To track rewards in A-E PN, we introduce a \textit{transitions reward function} \(\mathcal{R}\) that associates a reward to the firing of any transition, and we embed the total reward accumulated by firing transitions, which we call \textit{network reward} \(\rho\), in the network's marking.
In general, a reward can be produced by any change in the environment, regardless of whether an action or an evolution produced such change. For this reason, a reward is produced due to the firing of any transition, regardless if the transition is tagged as an action or an evolution. To comply with the classic RL cycle, rewards associated with evolutions are accumulated and awarded to the last action taken, eventually after a normalization operation (see subsection \ref{extending_rl_loop}).

\begin{figure}[h!]
    \centering
    \centerline{\includegraphics[width=\textwidth]{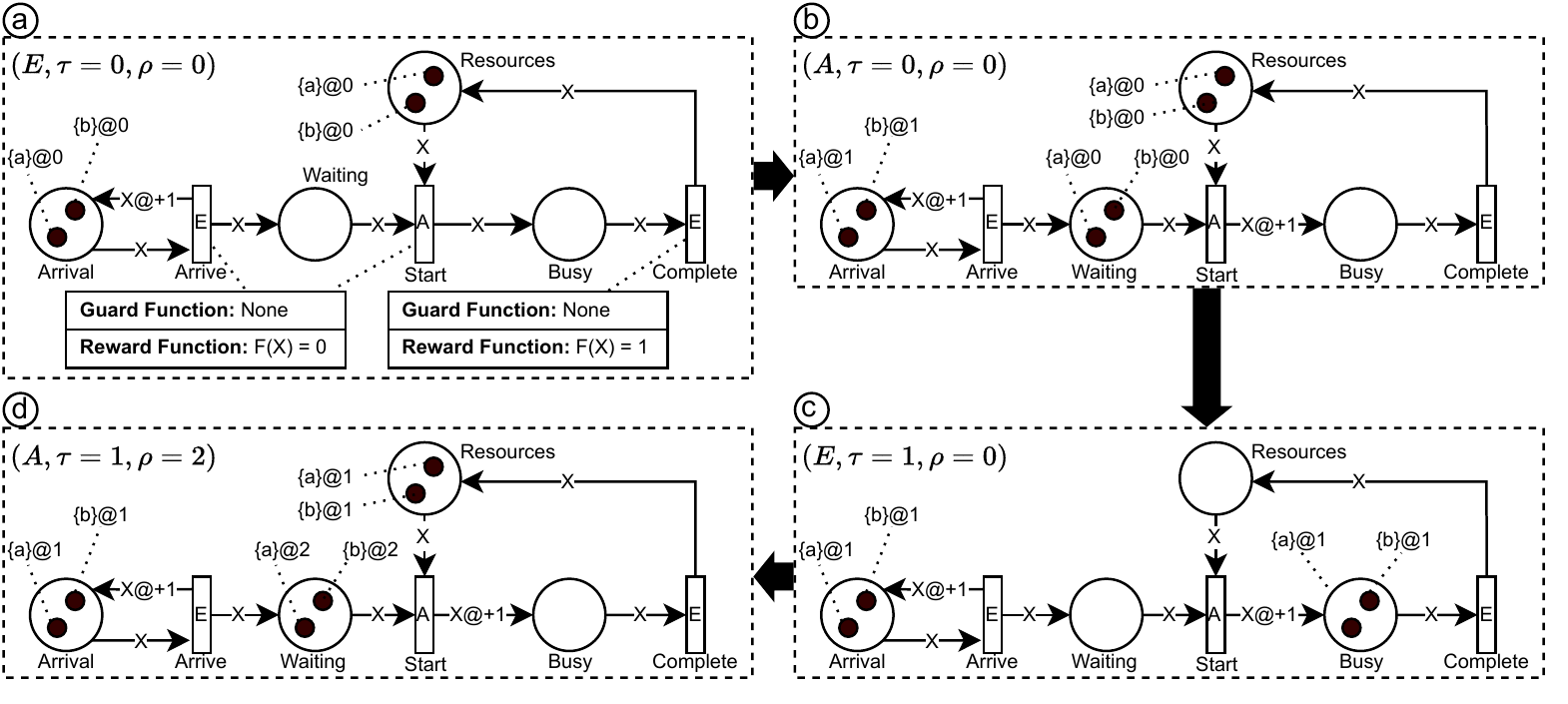}}
    \caption{A sequence of firings in a simple task assignment problem.}
    \label{fig:example_aepn}
\end{figure}

To further clarify the basic mechanisms of A-E PN, the example in ~\cref{fig:example_aepn} provides an overview of a sequence of firings.

The network shows the evolution of a system with two types of tasks, \(a\) and \(b\), and two employees, one that can undertake only task \(a\) and one that can undertake only task \(b\). A task of each type arrives at every clock tick, and an employee is assigned to a task of the same type. Assignments take one clock tick to complete, and a reward of \(1\) is produced every time an assignment is completed. The parentheses on the top right corner contain the components of the tagged marking that are not directly represented as network elements. Guard functions and reward functions are associated with single transitions. Timed tokens and arcs follow the notation introduced in \cref{def:tacpn}. The initial marking is presented in the dotted square \textit{a}, in which only \(E\) transitions are enabled. After two firings of transition \textit{Arrive}, consuming both tokens in the \textit{Arrival} place (in any order), no evolution transitions are available, so the tag is updated, and the system transitions to state \textit{b}. Notice that the transition from \textit{e} to \textit{a} does not produce a clock update, since actions are available to be taken at time \(0\). In \textit{b}, transition \textit{Start} is enabled. In this case, the RL agent would have two available actions: pairing task \(a\) with resource \(a\), or pairing task \(b\) with resource \(b\). In this case, both actions will be taken sequentially, in any order, leading to tagged marking \textit{c}, while in the general case, choices would have to be made by a decision algorithm on which assignments to make. In \textit{c}, the network tag is again \(E\), and two transitions are associated with time-enabled steps: \textit{Arrive} and \textit{Complete}. The firing of \textit{Arrive} produces two new tokens at time \(1\) in the \textit{Waiting} place, while the firing of \textit{Complete} places two tokens back in the \textit{Resources} place at time \(1\) and generates a network reward increment of \(2\) units in state \textit{d}.

\subsection{Formal Definition of Action-Evolution Petri Net}\label{formal_ae_pn}
To provide a formal definition of A-E PN, we must adapt three definitions from T-A CPN: the net itself, the marking, and the firing rules. 

\begin{sloppypar}
\begin{definition}[Action-Evolution Petri Net] Let \(\mathcal{T} = \{A, E\}\) be a finite \textit{set of tags} representing actions and evolutions, and \(S: \mathcal{T} \to \mathcal{T}\) a \textit{network tag} update function. An Action-Evolution Petri Net (A-E PN) is as a tuple \(AEPN =( \mathcal{E}, P, T, F, C, G, E, I, L, l_o, \mathcal{R}, \rho_0)\), where \( \mathcal{E}, P, T, F, C, G, E, I \) follow \cref{def:tacpn}, and:
\begin{itemize}
\item \(L : T \to \mathcal{T}\) is a \textit{transition tag function} that maps each transition \(t\) to a single tag. Only transitions associated with the same tag as the network can fire.
\item  \(l_0 \in \mathcal{T}\) is a singleton containing the \textit{network's initial tag}, usually equal to \(E\).
\item \(\mathcal{R} : T \to (f: \mathbb{R})\) associates every transition with a reward function. The function can take timing properties or numbers of tokens (representing completed cases) as parameters, thus allowing for flexbility in modeling reward.
\item \(\rho_0 \in \mathbb{R}\) is the initial \textit{network reward}, usually equal to \(0\).
\end{itemize}
\end{definition}
\end{sloppypar}

\begin{sloppypar}
\begin{definition}[Tagged Marking] A \textit{tagged marking} is a tuple \(TM = (M, l, \tau, \rho )\), where the tuple \((M, \tau)\) is a timed marking, as in \cref{def:tacpn_marking}, \(l \in \mathcal{T}\) is the \textit{network tag} at the current time \(\tau\), and \(\rho \in \mathbb{R}\) is the \textit{total reward} accumulated until the current time \(\tau\).
\end{definition}
\end{sloppypar}

\begin{sloppypar}
\begin{definition}[A-E PN Firing Rule]
\begin{enumerate}
\item A transition $t$ is tag-enabled in a tagged marking $(M, l, \tau, \rho)$ for binding $Y$ if and only if $t$ is enabled in \(M\) according to \cref{def:cpn}, and $L(t) = l$.   
\item Let $t$ be a transition that is tag-enabled in tagged marking $(M, l, \tau, \rho)$ for binding $Y = \langle v1=c_1@\tau_1, v2=c_2@\tau_2, \ldots, v_n=c_n@\tau_n \rangle$ . The enabling time of the transition, denoted $\tau_E$, is $max(\tau_1, \tau_2, \ldots, \tau_n)$.
\item An enabled transition $t$ is tag-time-enabled in tagged marking $TTM = (M, l, \tau, \rho)$, if its enabling time $\tau_E$ is less than or equal to $\tau$, and there exists no transition $t'$ that is enabled in tagged marking $TTM$ for some binding $Y'$ with enabling time $\tau'_E \leq \tau_E$.
\item A transition $t$ that is tag-time-enabled in tagged marking $(M, l, \tau, \rho)$ for binding $Y$ with enabling time $\tau_E$ can fire, changing the tagged marking to $(M', l, \tau_E, \rho')$, where $M'$ is constructed, such that $\forall p \in P: M'(p) = M(p) - E((p,t))_0\langle Y\rangle + E((t, p))_0\langle Y\rangle@\tau_E+E((t, p))_1$ and $\rho' = \rho + \mathcal{R}(t)$.
\item When there exists no $t$ in tagged marking $TTM = (M, l, \tau, \rho)$, for which there is a binding $Y$, such that $t$ is time-enabled, the set of all transitions is partitioned in two disjoint sets: \(T_{\text{current}} = \{t \in T | L(t) = l\}\) and \(T_{\text{next}} = \{t \in T | L(t) \neq l\}\). Let $\tau_{current}$ be the minimum value for which a transition in $T_{\text{current}}$ is time-enabled (according to \cref{def:tacpn_firing_rules}), and let $\tau_{next}$ be the minimum value for which a transition in $T_{\text{next}}$ is time-enabled. Note that $\tau_{current}$ and $\tau_{next}$ can be undefined.
\begin{itemize}
\item If $\tau_{current}$ is defined, and $\tau_{current} \leq \tau_{next}$ or $\tau_{next}$ is undefined, only the global clock is updated, leading to a new tagged marking $TTM' = (M, l, \tau_{current}, \rho)$.
\item If $\tau_{next}$ is defined, and $\tau_{current} > \tau_{next}$ or $\tau_{current}$ is undefined, both the global clock and the network tag are updated, leading to a new tagged marking $TTM' = (M, S(l), \tau_{next}, \rho)$.
\end{itemize}

\end{enumerate}
\end{definition}
\end{sloppypar}

\subsection{Extending the Reinforcement Learning Loop}\label{extending_rl_loop}

\begin{figure}[h!]
    \centering
    \centerline{\includegraphics[width=\textwidth]{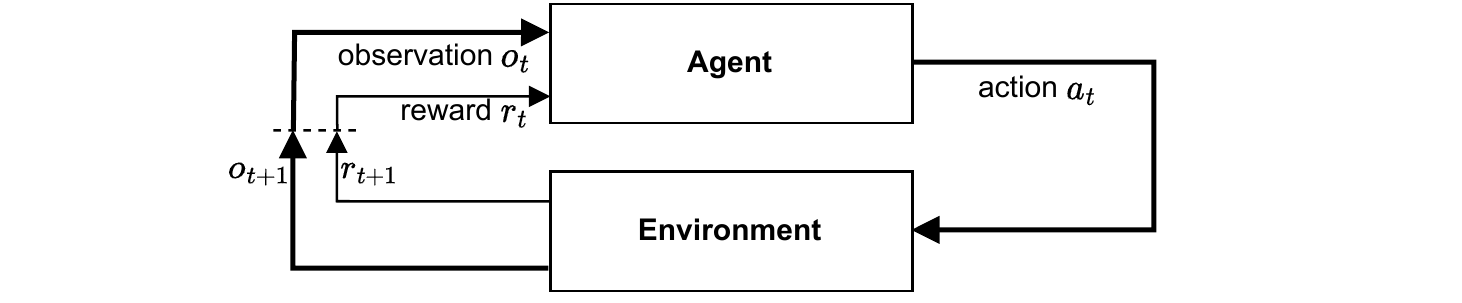}}
    \caption{A common representation of the RL training cycle~\cite{sutton_reinforcement_nodate}.}
    \label{fig:rl}
\end{figure}

Having completely defined the characteristics of the A-P PN formalism, we can clarify how it can be used to learn optimal task assignment policies (i.e. mapping from observations to assignments) by applying it in a Reinforcement Learning (RL) cycle. Figure~\ref{fig:rl} shows the RL cycle. In every step in the cycle, the agent receives an observation (a representation of the environment's state), then it produces a single action that it considers the best action for this observation. The action leads to a change in the environment's state. The environment is responsible for providing a reward for the chosen action along with a new observation. Then the cycle repeats, and a new decision step takes place. The MDP formulation is the standard framework for training an agent to take actions that lead to the highest cumulative reward. 

\begin{figure}[]
    \centering
    \centerline{\includegraphics[width=0.9\textwidth]{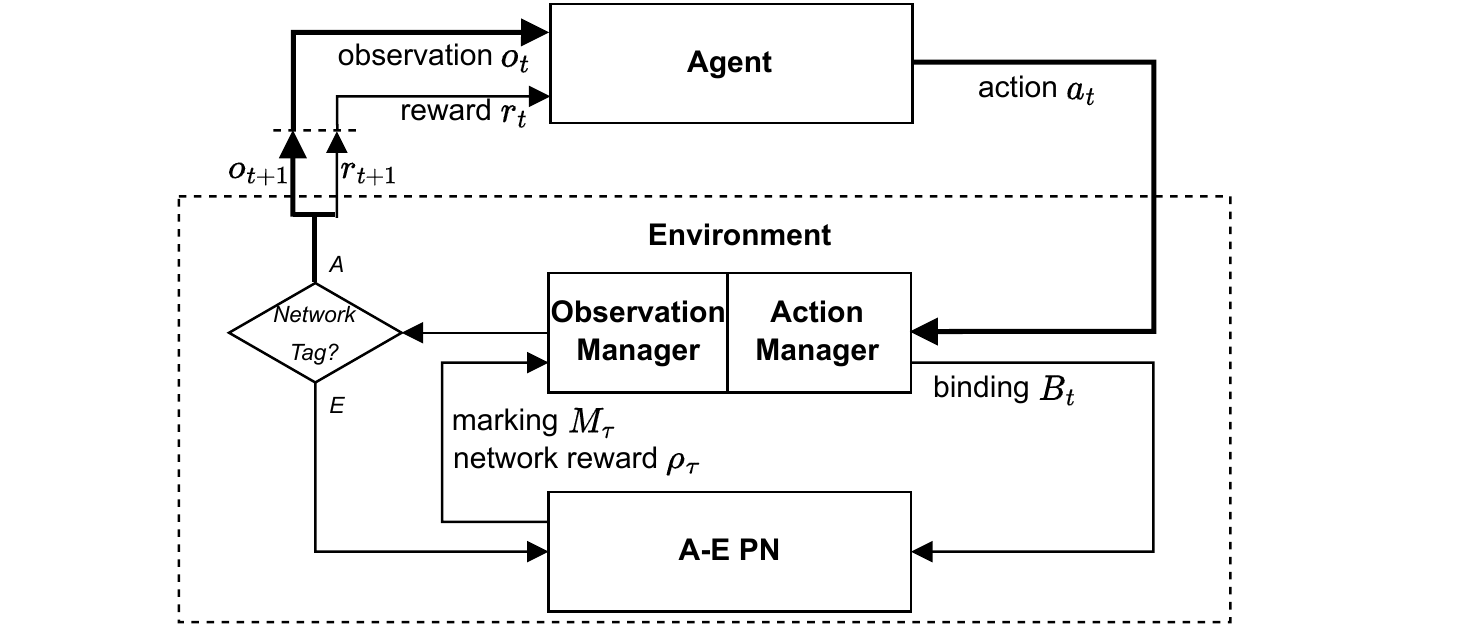}}
    \caption{The reinforcement learning cycle with A-E PN}
    \label{fig:rl_extended}
\end{figure}

In recent years, the embedding of neural networks in RL algorithms gave birth to the field of Deep Reinforcement Learning (DRL), achieving breakthroughs in settings such as playing board games~\cite{silver_mastering_2017} and robotic manipulation~\cite{kalashnikov_qt-opt_2018}, as well as successful applications in domains like industrial process control~\cite{nian_review_2020}, and healthcare~\cite{yu_reinforcement_2020}. With the proliferation of robust DRL algorithms, the main hurdle in modeling new problems is the definition of the environment, which is usually represented as a black box, as in \cref{fig:rl}, thus leaving the implementation of the system's dynamics entirely to the modeler. The lack of a standardized interface makes the creation of new environments time-consuming and dependent on the modeler's coding skills. Moreover, even introducing small changes potentially requires substantial effort once the environment has been modeled. These observations motivate the effort to provide a unified and executable framework. In \cref{fig:rl_extended}, the classic RL cycle is extended to account for the presence of A-E PN. The main element is the A-E PN, which acts as a simulator for the whole process. 

The A-E PN communicates with the agent through two sub-components: \textit{observation manager} and \textit{action manager}. The observation manager is invoked every time the tagged marking changes, regardless if due to a firing or not. The new reward is stored, and the network tag is evaluated: if the tag is \(E\), no action is required, and the control is given back to the A-E PN, which can fire a new \(E\) transition. If the tag is \(A\), the accumulated rewards are added up, and the result is divided by \(1 + (\tau _{t+1}- \tau _{t})\). The resulting value is returned to the agent as \(r_{t+1}\). The reward value takes into account the possible misalignment between clock ticks (\(\tau\)) and RL steps (\(t\)), given by the fact that multiple actions can happen at the same \(\tau\). The observation manager also returns to the agent the new observation \(o_{t+1}\). For the set of experiments presented in the next section, the observation is built as a vector containing, for each place, the number of tokens of each color in the place's color set. The action manager is invoked every time the agent chooses an action \(a_t\), which it transforms into the corresponding binding \(B_t\) (associated with an action transition) to be fired.

\section{Evaluation}\label{use_cases}
This section aims to show that A-E PN constitutes a unified and executable framework for expressing dynamic task assignment problems with different characteristics: in fact, all the examples were modeled using a single notation (except for color-specific functions on arcs, guards, and rewards) and a RL algorithm was trained on each problem, without any additional development effort.

We provide a (non-exhaustive) taxonomy of assignment problem variants based on \cite{pentico_assignment_2007}. We distinguish three archetypes of assignment problems.
 \begin{itemize}
     \item \textbf{Assignment Problem with Compatibilities}: resources are assigned to tasks according to a measure of compatibility. Two problem subclasses can be formulated:
     \begin{itemize}
      \item \textbf{Assignment Problem with Hard Compatibilities}: resources can only be assigned to tasks if they are compatible. The dynamic task assignment problem in subsection \ref{dynamic_task_assignment} falls into this subclass.
      \item \textbf{Assignment Problem with Soft Compatibilities}: resources can always be assigned to tasks, but different assignments result in different system behaviors. An example of such a problem is if multiple resources can perform a task, but some will be faster at it than others.
     \end{itemize}
     \item \textbf{Assignment Problem with Multiple Assignments}: the same resource can be assigned to multiple tasks, or the same task can be assigned to 
     multiple resources. Two problem subclasses can be formulated:
     \begin{itemize}
     \item \textbf{Assignment Problem with Resource Capacity}: resources have a maximum capacity of tasks that they can undertake before being considered full. In the simple case each resource can only be busy with a single task at a time. The dynamic bin packing problem in subsection \ref{bin_packing} provides a more elaborate example.
     \item \textbf{Assignment Problem with Task Capacity}: tasks have a minimum capacity of resources to be assigned to them before processing. In the simple case each tasks needs exactly one resource.
     \end{itemize}
     \item \textbf{Assignment Problem with Dynamic Resources' Behavior}: resources have dynamic behavior. Two problem subclasses can be formulated:
     \begin{itemize}
     \item \textbf{Assignment Problem with Action-Dependent Dynamic Resources' Behavior}: resources change their attribute values as the consequence of taking actions. The dynamic order-picking problem in subsection \ref{order_picking} falls into this category.
     \item \textbf{Assignment Problem with Action-Independent Dynamic Resources' Behavior}: resources change their attribute values as the consequence of evolutions in the environment. For example, resources may take breaks or go on holidays.
     \end{itemize}
\end{itemize}
In the following sections, one example is detailed for each archetype. An example for each subclass is implemented in the provided Python package.

\subsection{Dynamic Task Assignment Problem with Hard Compatibilities}\label{dynamic_task_assignment}
\begin{figure}[]
    \centering
    \centerline{\includegraphics[width=\textwidth]{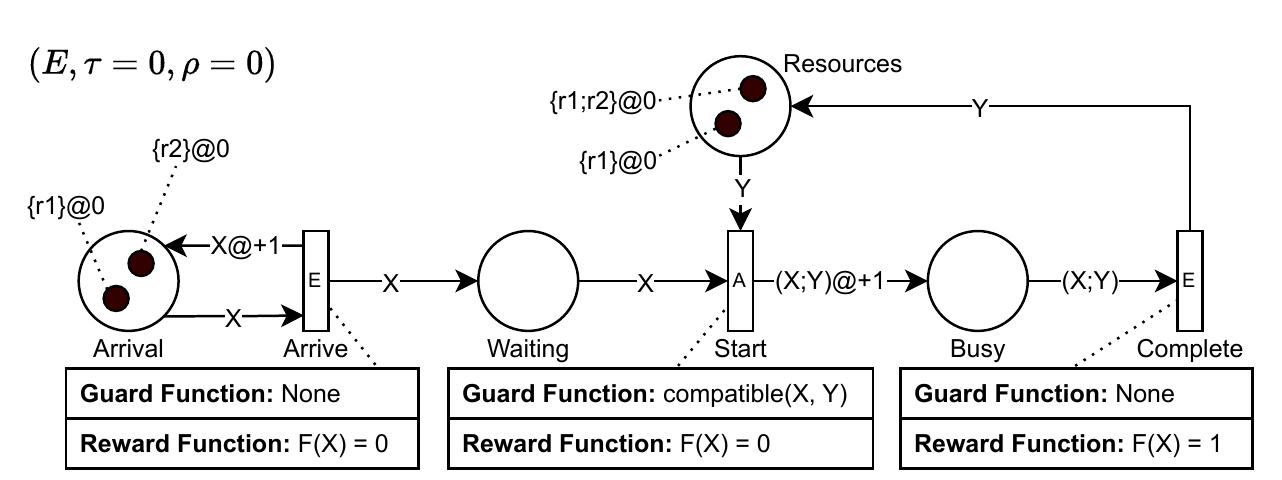}}
    \caption{A-E PN initial marking for the dynamic task assignment problem}
    \label{fig:task_assignment}
\end{figure}

Let us consider a system that solves a task assignment problem, similar to the one presented in \cref{fig:example_aepn}. At every clock tick, two tasks arrive: one has type \(r1\) and the other \(r2\). Two resources are available for the assignment: one can only undertake tasks of type \(r1\), while the other can undertake tasks of type \(r1\) or \(r2\).
Once a task is assigned to a resource, completion always takes one clock tick, after which the resource becomes available for a new assignment. A resource cannot work on multiple tasks at the same time. A network reward of \(1\) is returned every time a task is assigned to a resource and every time an assignment completes, leading to a theoretical maximum reward of \(200\) over \(100\) clock ticks. The problem can be fully expressed in terms of A-E PN, as reported in \cref{fig:task_assignment}.

\subsection{Dynamic Bin Packing Problem}\label{bin_packing}
\begin{figure}[h!]
    \centering
    \centerline{\includegraphics[width=\textwidth]{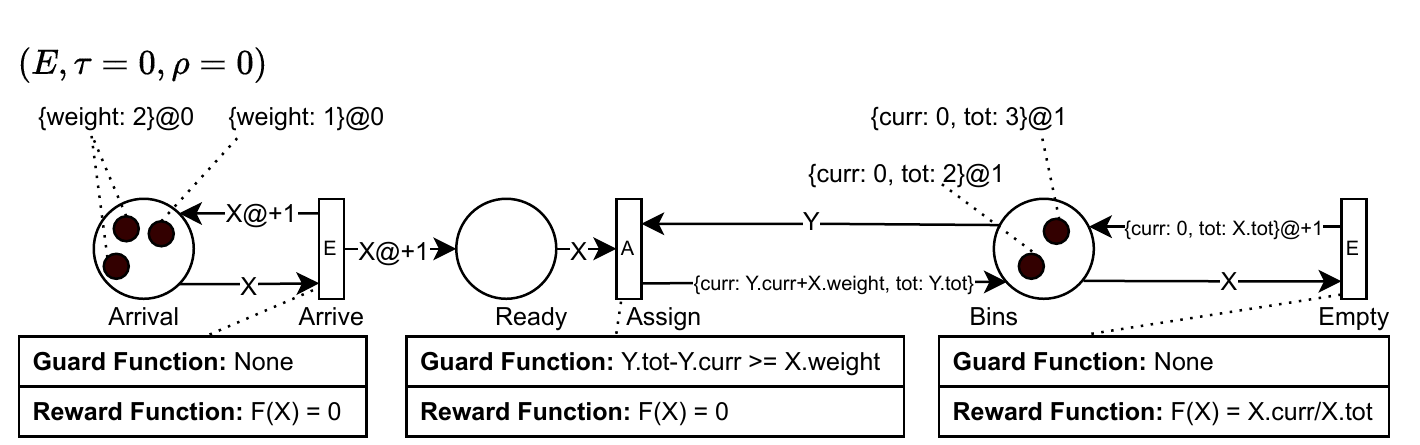}}
    \caption{A-E PN initial marking for the dynamic bin packing problem}
    \label{fig:bin_packing}
\end{figure}

In this scenario, we model a dynamic version of the bin packing problem where items (the problem tasks, characterized by their \textit{weight}) arrive sequentially and they must be allocated to two bins (the problem resources, characterized by the total weight of objects in the bin \textit{curr} and the bin's total capacity \textit{tot}) that are emptied at every clock tick (except for the first, which is used to generate the objects to be put in the bins). The fullness of the bins before being emptied gives the measure of goodness of the object's allocation, quantified as the weight of objects in the bin divided by the total bin capacity. This problem showcases how tokens' colors can be used to model non-trivial reward functions. In the example reported, three objects arrive in the system at every clock tick, one of weight \(1\) and two of weight \(2\). Two initially empty bins are available, one with capacity \(2\) and one with capacity \(3\). The optimal allocation would give a reward of \(2\), leading to a theoretical maximum reward of \(200\) over a \(100\) clock ticks horizon. The A-E CPN formalization of the problem is reported in \cref{fig:bin_packing}.

\subsection{Dynamic Order-Picking Problem}\label{order_picking}

\begin{figure}[h!]
    \centering
    \centerline{\includegraphics[width=\textwidth]{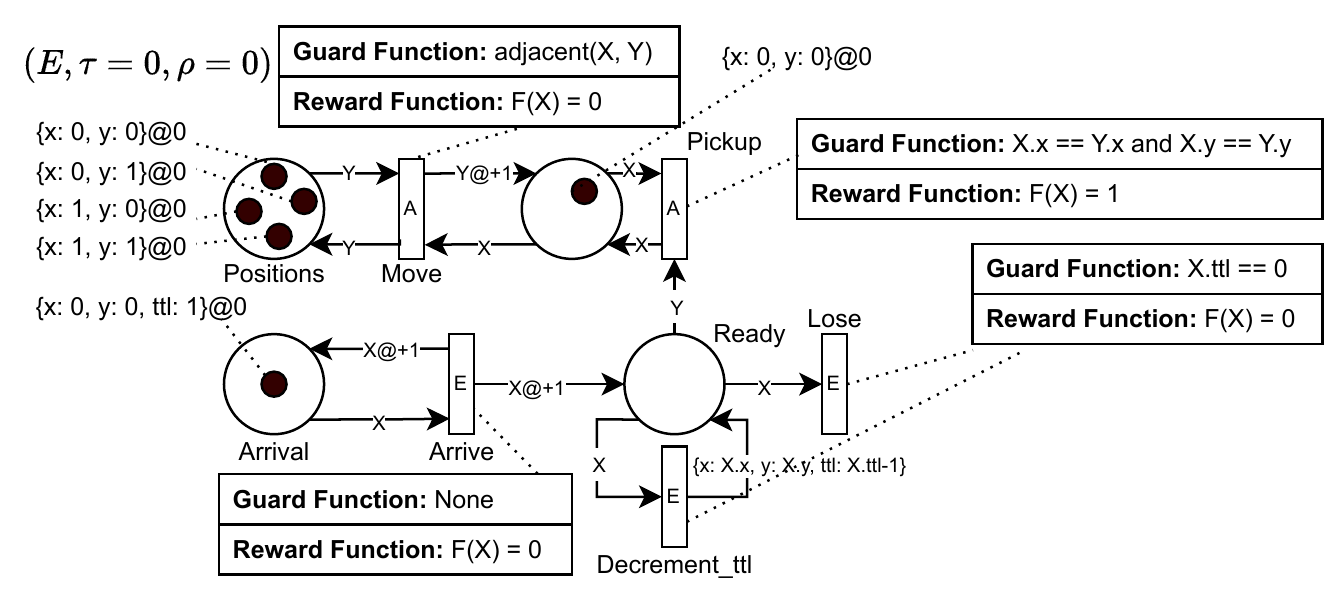}}
    \caption{A-E PN initial marking for the dynamic order-picking problem}
    \label{fig:vrp}
\end{figure}

In this section, we present an example of action-dependent resource behavior (i.e. the agent taking decisions on the actions that it performs). The example is a simple order-picking problem in which a single agent (the resource) moves on a squared grid of size \(2\), trying to pick orders (the tasks). The agent's and the orders' colors are characterized by two parameters representing the coordinates on the grid (infinite capacity is assumed). The agent starts in position \((0,0)\) and can move left, right, up, or down, but not over a diagonal. If an order is in the same position as the agent, the latter can use an action to pick the order. A single order arrives at every clock tick, always in position \(1,1\), and the order stays on the grid for exactly one clock tick, according to a time-to-live (TTL) parameter. The agent's objective is to pick as many orders as possible, so it gets a reward of \(1\) every time an order is picked, leading to a theoretical maximum reward of \(98\) over a \(100\) clock ticks horizon (at least two orders will be lost due to the agent moving to position \((1,1)\). The problem is formulated in terms of A-E PN in \cref{fig:vrp}.

\subsection{Experimental Results}\label{results}

All experiments were implemented in a proof-of-concept package\footnote{The code is publicly 
available in \url{https://github.com/bpogroup/aepn-project}.}, relying on the Python programming language and the widely adopted RL library Gymnasium~\cite{gymnasium}. Proximal Policy Optimization (PPO)~\cite{schulman_proximal_2017} with masking was used as the training algorithm. Specifically, the PPO implementation of the \textit{Stable Baselines} package~\cite{sb_contrib} is used. Note, however, that the mapping from each A-E PN to PPO was automated and requires no further effort from the modeler. The PPO algorithm was trained on each example for (\(10^6\) steps with \(100\) clock ticks per episode, completed in less than \(2300\) seconds on a mid-range laptop, without GPUs), always using the default hyperparameters. The experimental results were computed on (network) rewards obtained by the trained agent and following a random policy over \(1000\) trajectories, each of duration \(100\) clock ticks.
In \cref{tab:results}, the average and standard deviations of rewards obtained by the trained PPO are compared to those of a random policy on each of the three presented problem instances, with reference to the maximum attainable reward. In all cases, PPO shows to be able to learn a close-to-optimal assignment policy.

\setlength{\tabcolsep}{5pt}
\begin{table}[]
\caption{The results for the three presented problem instances.}
\label{tab:results}
\centering
\begin{tabular}{llll}
\toprule
Instance        & Random                & PPO                   & Optimal \\ \midrule
Task Assignment & \(186.894 \pm 2.084\)  & \(199.852 \pm 0.398\)  & \(200\) \\
Bin Packing     & \(186.746 \pm 1.941\) & \(199.963 \pm 0.186\) & \(200\) \\
Order Picking   & \(6.046 \pm 2.585\)   & \(96.776 \pm 2.019\)  & \(98\)  \\ \bottomrule
\end{tabular}
\end{table}
\section{Conclusions and Future Work}\label{conclusion}

This paper presented a framework for modeling and solving dynamic task assignment problems. To this end, it introduced a new variant of Petri Nets, namely Action-Evolution Petri Nets (A-E PN), to provide a mathematically sound modeling tool. This formalism was integrated with the Reinforcement Learning (RL) cycle and consequently with existing algorithms that can solve RL problems. To evaluate the general applicability of the framework for modeling and solving task assignment problems, a taxonomy of archetypical problems was introduced, and working examples were provided. A DRL algorithm was trained on each implementation, obtaining close-to-optimal policies for each example. This result shows the suitability of A-E PN as a unified and executable framework for modeling and solving assignment problems. 

While the applicability of the framework was shown, its possibilities and limitations are yet to be fully explored. This will be done in future research by expanding the provided taxonomy of assignment problems and considering different problem classes.
\section*{Acknowledgement}
The research that led to this publication was partly funded by the European Supply Chain Forum (ESCF) and the Eindhoven Artificial Intelligence Systems Institute (EAISI) under the AI Planners of the Future program.

\bibliographystyle{unsrt}
\bibliography{lnbib} 

\begin{thebibliography}{10}

\bibitem{kuhn_hungarian_1955}
H.~W. Kuhn.
\newblock The {Hungarian} method for the assignment problem.
\newblock {\em Naval Research Logistics Quarterly}, 2(1-2):83--97, March 1955.

\bibitem{gulpinar_heuristics_2018}
N.~Gülpınar, E.~Çanakoğlu, and J.~Branke.
\newblock Heuristics for the stochastic dynamic task-resource allocation
  problem with retry opportunities.
\newblock {\em European Journal of Operational Research}, 266(1):291--303,
  April 2018.

\bibitem{hu_petri-net-based_2020}
L.~Hu, Z.~Liu, W.~Hu, Y.~Wang, and J.~Tan.
\newblock Petri-net-based dynamic scheduling of flexible manufacturing system
  via deep reinforcement learning with graph convolutional network.
\newblock {\em Journal of Manufacturing Systems}, 55:1--14, April 2020.

\bibitem{spivey_dynamic_2004}
M.~Z. Spivey and W.~B. Powell.
\newblock The {Dynamic} {Assignment} {Problem}.
\newblock {\em Transportation Science}, 38(4):399--419, November 2004.

\bibitem{sutton_reinforcement_nodate}
R.~Sutton and A.~Barto.
\newblock Reinforcement {Learning}: {An} {Introduction}.
\newblock page 352.

\bibitem{goos_brief_1997}
K.~Jensen.
\newblock A brief introduction to coloured {Petri} {Nets}.
\newblock In {\em Tools and {Algorithms} for the {Construction} and {Analysis}
  of {Systems}}, volume 1217, pages 203--208. 1997.

\bibitem{bause_stochastic_1998}
F.~Bause and P.~Kritzinger.
\newblock Stochastic {Petri} {Nets}: {An} {Introduction} to the {Theory}.
\newblock {\em ACM SIGMETRICS Performance Evaluation Review}, 26(2):2--3,
  August 1998.

\bibitem{da_rocha_costa_markov_2010}
M.~Eboli and F.~Cozman.
\newblock Markov {Decision} {Processes} from {Colored} {Petri} {Nets}.
\newblock In {\em Advances in {Artificial} {Intelligence} – {SBIA} 2010},
  volume 6404, pages 72--81. 2010.
\newblock Series Title: Lecture Notes in Computer Science.

\bibitem{hutchison_markov_2007}
M.~Beccuti, G.~Franceschinis, and S.~Haddad.
\newblock Markov {Decision} {Petri} {Net} and {Markov} {Decision}
  {Well}-{Formed} {Net} {Formalisms}.
\newblock In {\em Petri {Nets} and {Other} {Models} of {Concurrency} –
  {ICATPN} 2007}, volume 4546, pages 43--62. 2007.

\bibitem{qiu_dynamic_2000}
Q.~Qiu, Q.~Wu, and M.~Pedram.
\newblock Dynamic power management of complex systems using generalized
  stochastic {Petri} nets.
\newblock In {\em Proceedings of the 37th conference on {Design} automation -
  {DAC} '00}, pages 352--356. ACM Press, 2000.

\bibitem{drakaki_manufacturing_2017}
M.~Drakaki and P.~Tzionas.
\newblock Manufacturing {Scheduling} {Using} {Colored} {Petri} {Nets} and
  {Reinforcement} {Learning}.
\newblock {\em Applied Sciences}, 7(2):136, February 2017.

\bibitem{riedmann_timed_2022}
S.~Riedmann, J.~Harb, and S.~Hoher.
\newblock Timed {Coloured} {Petri} {Net} {Simulation} {Model} for
  {Reinforcement} {Learning} in the {Context} of {Production} {Systems}.
\newblock In Bernd-Arno Behrens, Alexander Brosius, Welf-Guntram Drossel,
  Wolfgang Hintze, Steffen Ihlenfeldt, and Peter Nyhuis, editors, {\em
  Production at the {Leading} {Edge} of {Technology}}, pages 457--465, Cham,
  2022. Springer International Publishing.

\bibitem{jensen_high-level_1991}
K.~Jensen and G.~Rozenberg.
\newblock {\em High-level {Petri} nets: theory and application}.
\newblock Springer-Verlag, 1991.

\bibitem{10.1007/978-3-642-18381-2_4}
L.~Jacobsen, M.~Jacobsen, M.~H. M{\o}ller, and J.~Srba.
\newblock Verification of timed-arc petri nets.
\newblock In {\em SOFSEM 2011: Theory and Practice of Computer Science}, pages
  46--72, 2011.

\bibitem{cpntools}
{CPN Tools}.
\newblock \url{https://cpntools.org/}.

\bibitem{silver_mastering_2017}
D.~Silver, J.~Schrittwieser, K.~Simonyan, I.~Antonoglou, A.~Huang, A.~Guez,
  T.~Hubert, L.~Baker, M.~Lai, A.~Bolton, Y.~Chen, T.~Lillicrap, F.~Hui,
  L.~Sifre, G.~van~den Driessche, T.~Graepel, and D.~Hassabis.
\newblock Mastering the game of {Go} without human knowledge.
\newblock {\em Nature}, 550(7676):354--359, October 2017.

\bibitem{kalashnikov_qt-opt_2018}
D.~Kalashnikov, A.~Irpan, P.~Pastor, J.~Ibarz, A.~Herzog, E.~Jang, D.~Quillen,
  E.~Holly, M.~Kalakrishnan, V.~Vanhoucke, and S.~Levine.
\newblock {QT}-{Opt}: {Scalable} {Deep} {Reinforcement} {Learning} for
  {Vision}-{Based} {Robotic} {Manipulation}, November 2018.

\bibitem{nian_review_2020}
R.~Nian, J.~Liu, and B.~Huang.
\newblock A review on reinforcement learning: Introduction and applications in
  industrial process control.
\newblock {\em Computers \& Chemical Engineering}, 139:106886, August 2020.

\bibitem{yu_reinforcement_2020}
C.~Yu, J.~Liu, and S.~Nemati.
\newblock Reinforcement {Learning} in {Healthcare}: {A} {Survey}, April 2020.
\newblock arXiv:1908.08796 [cs].

\bibitem{pentico_assignment_2007}
David~W. Pentico.
\newblock Assignment problems: {A} golden anniversary survey.
\newblock {\em European Journal of Operational Research}, 176(2):774--793,
  January 2007.

\bibitem{gymnasium}
{Gymnasium}.
\newblock \url{https://gymnasium.farama.org/}.

\bibitem{schulman_proximal_2017}
J.~Schulman, F.~Wolski, P.~Dhariwal, A.~Radford, and O.~Klimov.
\newblock Proximal {Policy} {Optimization} {Algorithms}, August 2017.
\newblock arXiv:1707.06347 [cs].

\bibitem{sb_contrib}
Sb3-contr.
\newblock
  \url{https://github.com/Stable-Baselines-Team/stable-baselines3-contrib}.

\end{thebibliography}

\end{document}